\pdfoutput=1
\documentclass{article}

    \usepackage[preprint, nonatbib]{neurips_2021}

\usepackage{tikz} 

\usepackage{multicol}
\usepackage{bm}

\usepackage[most]{tcolorbox}
\usepackage{subfig}

\newcommand{\mysubfig}[3][width=\linewidth]{%
    \tcbitem\subfloat[#2]{\includegraphics[#1]{#3}}}

\usepackage{mwe}
\usepackage{caption} 

\usepackage{floatrow}
\usepackage{adjustbox}

\usepackage{tabularx}
\usepackage{geometry}
\usepackage{lipsum}
\usepackage{graphicx}

\usepackage[utf8]{inputenc} 
\usepackage[T1]{fontenc}    
\usepackage{hyperref}       
\usepackage{url}            
\usepackage{booktabs}       
\usepackage{amsfonts}       
\usepackage{nicefrac}       
\usepackage{microtype}      
\usepackage{float}          
\usepackage{colortbl}         
\usepackage[normalem]{ulem} 

\usepackage{multirow}
\usepackage{siunitx}
\usepackage{xspace}

\newcommand{\lossqd}{\mathcal{L}_{PI}}
\newcommand{\MethodName}{PIVEN\xspace}
\newcommand{\Vhalf}{MOI\xspace}
\newcommand{\OnlyRmse}{POO\xspace}

\newcommand{\RR}{\mathbb{R}} 

\usepackage{amssymb}
\usepackage{amsthm}
\usepackage{enumitem}
\usepackage{mathtools}
\newcommand{\defeq}{\vcentcolon=} 
\usepackage{amsmath}
\DeclareMathOperator*{\argmin}{arg\,min} 

\definecolor{Gray}{gray}{0.85}
\newcolumntype{a}{>{\columncolor{Gray}}c}

\title{PIVEN: A Deep Neural Network for Prediction Intervals with Specific Value Prediction}

\author{%
  Eli Simhayev \\
  Ben-Gurion University of the Negev\\
  Beer-Sheva, Israel \\
  \texttt{elisim@post.bgu.ac.il} \\
   \And
   Gilad Katz \\
   Ben-Gurion University of the Negev \\
    Beer-Sheva, Israel \\
   \texttt{giladkz@post.bgu.ac.il} \\
    \And
   Lior Rokach \\
   Ben-Gurion University of the Negev \\
    Beer-Sheva, Israel \\
   \texttt{liorrk@post.bgu.ac.il} \\
}

\begin{document}

\maketitle

\begin{abstract}
Improving the robustness of neural nets in regression tasks is key to their application in multiple domains. Deep learning-based approaches aim to achieve this goal either by improving their prediction of specific values (i.e., point prediction), or by producing prediction intervals (PIs) that quantify uncertainty. We present \MethodName, a deep neural network  for producing both a PI and a value prediction. Our loss function expresses the value prediction as a function of the upper and lower bounds, thus ensuring that it falls within the interval without increasing model complexity. Moreover, our approach makes no assumptions regarding data distribution within the PI, making its value prediction more effective for various real-world problems. Experiments and ablation tests on known benchmarks show that our approach produces tighter uncertainty bounds than the current state-of-the-art approaches for producing PIs, while maintaining comparable performance to the state-of-the-art approach for value-prediction. Additionally, we go beyond previous work and include large image datasets in our evaluation, where \MethodName is combined with modern neural nets.
\end{abstract}

\section{Introduction}

Deep neural networks (DNNs) have been achieving state-of-the-art results in a large variety of complex problems. These include automated decision making and recommendation in the medical domain \cite{medical_dnn}, autonomous drones \cite{dronednn} and self driving cars \cite{carsdnn}. In many of these domains, it is crucial not only that the prediction made by the DNN is accurate, but rather that its uncertainty is quantified. Quantifying uncertainty has many benefits, including risk reduction and more reliable planning \cite{lube}. In regression, uncertainty is quantified using prediction intervals (PIs), which offer upper and lower bounds on the value of a data point for a given probability (e.g., 95\%). Existing non-Bayesian PI generation methods can be roughly divided into two groups: \textit{a)} performing multiple runs of the regression problem, as in dropout \cite{mc_dropout} or ensemble-based methods \cite{deep_ensemble}, then deriving post-hoc the PI from prediction variance, and; \textit{b)} dedicated architectures for the PI generation \cite{qd, SQR, salem}. 

While effective, both approaches have shortcomings. The former group is not optimized for PIs generation, having to convert a set of sampled values into a distribution. This lack of PI optimization makes using these approaches difficult in domains such as financial risk mitigation or scheduling. For example, providing a PI for the number of days a machine can function without malfunctioning (e.g., 30-45 days with 99\% certainty) is more valuable than a prediction for the specific time of failure. 

The latter group---PI-dedicated architectures---provides accurate upper and lower bounds for the prediction, but finds it difficult to produce accurate value predictions. One approach for producing values is to select the middle of the interval \cite{qd, SQR}, but this often leads to sub-optimal results due to various implicit assumptions regarding data distribution within the PI. Another recent approach \cite{salem} attempts to learn the value prediction simultaneously by adding terms to the loss function. While effective, the additional terms increase model complexity, consequently requiring heavy optimization.

We propose \MethodName (\textbf{p}rediction \textbf{i}ntervals with specific \textbf{v}alue pr\textbf{e}dictio\textbf{n}), a novel approach for simultaneous PI generation and value prediction using DNNs. Our approach combines the benefits of the two above-mentioned groups by producing \textit{both} a PI and a value prediction, while ensuring that the latter is within the former. By doing so, \MethodName ensures prediction integrity, without increasing model complexity.
We follow the experimental procedure of recent works, and compare our approach to current best-performing methods: \textit{a)} Quality-Driven PI (QD) \cite{qd}, a dedicated PI generation method; \textit{b)} QD-Plus \cite{salem}, a recent improvement of QD which adds additional terms for the value prediction, and; \textit{c)} Deep Ensembles (DE) \cite{deep_ensemble}. Our results show that \MethodName outperforms QD and QD-Plus by producing narrower PIs, while simultaneously achieving comparable results to DE in terms of value prediction. 
We are also the first to include large image datasets and explore the integration of our PI-producing methods with large convolutional architecture.

\section{Related Work}

Modeling uncertainty in deep learning has been an active area of research in recent years 
\cite{qd, rio, mc_dropout, deep_ensemble, keren_pi_calibrated, vision_uncertainties, bias_reduced, can_you_trust, zhu2017deep}. Studies in uncertainty modeling and regression can be generally divided into two groups: \textit{PI-based} and \textit{non-PI-based}. 
Non-PI approaches utilize both Bayesian \cite{bnn} and non-Bayesian approaches. The former methods define a prior distribution on the weights and biases of a neural net (NN), while inferring a posterior distribution from the training data. Non-Bayesian methods \cite{mc_dropout, deep_ensemble, rio} do not use initial prior distributions. In \cite{mc_dropout}, Monte Carlo sampling was used to estimate the predictive uncertainty of NNs through the use of dropout over multiple runs. A later study \cite{deep_ensemble} employed a combination of ensemble learning and adversarial training to quantify model uncertainty. In an expansion of a previously-proposed approach \cite{mve}, each NN was optimized to learn the mean and variance of the data, assuming a Gaussian distribution. Recently, \cite{rio} proposed a post-hoc procedure using Gaussian processes to measure uncertainty.

PI-based approaches are designed to produce a PI for each sample. \cite{keren_pi_calibrated} propose a post-processing approach that considers the regression problem as one of classification, and uses the output of the final softmax layer to produce PIs. \cite{SQR} propose the use of a loss function designed to learn all conditional quantiles of a given target variable. LUBE \cite{lube} consists of a loss function optimized for the creation of PIs, but has the caveat of not being able to use stochastic gradient descent (SGD) for its optimization. A recent study \cite{qd} inspired by LUBE, proposed a loss function that is both optimized for the generation of PIs and can be optimized using SGD. Recently, \cite{salem} proposed a method for combining PI generation and value prediction, but the approach requires computationally-heavy post-hoc optimization.

The two groups presented above tends to under-perform when applied to tasks for which its loss function was not optimized: Non-PI approaches produce more accurate value predictions, but are not optimized to produce PI and therefore produce bounds that are less tight. PI-based methods produce tight bounds, but tend to underperform when producing value predictions. Several recent studies attempted to produce both value predictions and PIs: \cite{adaptive, conformal_prediction_nips19} do so by using conformal prediction with quantile regression, while QD+ \cite{salem} simultaneously outputs a PI and a value prediction using an additional output. While effective, these approaches require either a complex splitting strategy \cite{adaptive, conformal_prediction_nips19}, or the addition of multiple terms to the loss function, consequently increasing model complexity and requiring computationally heavy post-hoc optimization. 
Contrary to these approaches, \MethodName produces PIs with value predictions by using a novel loss function and without any increases in complexity.

\section{Motivation}

\subsection{Problem Formulation}
\label{sec:problemFormulation}
We consider a NN regressor that processes an input \(x \in \mathcal{X}\) with an associated label \(y \in \RR\), where \({\mathcal{X}}\) can be any feature space (e.g., tabular data, age prediction from images). Let \((x_i,y_i) \in \mathcal{X} \times \RR \) be a data point along with its target value. Let \(U_i\) and \(L_i\) be the  upper and lower bounds of PIs corresponding to the ith sample. Our goal is to construct \((L_i, U_i, y_i)\) so that \(\texttt{Pr}(L_i \le y_{i} \le U_i) \ge 1-\alpha \). We refer to \(1-\alpha\) as the confidence level of the PI. 

We define two quantitative measures (Eq. \ref{eq:picp} and \ref{eq:mpiw}) for PIs evaluation (see \cite{lube}). When combined, these metrics enable us to comprehensively evaluate the quality of generated PIs. \textit{Coverage} is the ratio of dataset samples that fall within their respective PIs, measured using the \textit{prediction interval coverage probability} (PICP) metric. The variable $n$ denotes the number of samples and \(k_i = 1\) if \(y_i \in (L_i, U_i)\), otherwise \(k_i = 0\). \textit{Mean prediction interval width} (MPIW) is a quality metric for the generated PIs whose goal is producing as tight a bound as possible.

\noindent\begin{minipage}{.5\linewidth}
\begin{equation}\label{eq:picp}
  PICP \defeq \frac{1}{n}\sum_{i=1}^{n}
\end{equation}
\end{minipage}%
\begin{minipage}{.5\linewidth}
\begin{equation}\label{eq:mpiw}
  MPIW \defeq \frac{1}{n}\sum_{i=1}^{n} U_i-L_i
\end{equation}
\end{minipage}




\subsection{Motivation}
\label{subsec:motivation}
PI-producing NNs aim to minimize PI width (MPIW) while maintaining  predefined coverage (PICP). This approach has two significant drawbacks:

\noindent \textbf{Lack of value prediction capabilities}. Current PI-producing architectures do not output specific values (their original task). Most studies \cite{qd, SQR} circumvent this problem by returning the middle of the PI, but this approach leads to sub-optimal results for skewed distributions. We are aware of only one previous study that aims to address this challenge \cite{salem}, but it has shortcomings we discuss in our evaluation in Section \ref{subsec:evaluation}.

\noindent \textbf{Overfitting.} The MPIW metric is optimized for the predefined percentage of samples that fall within their generated PIs (defined by PICP). As a result, the NN essentially overfits to a subset of the data. 

\MethodName is specifically designed to address these shortcomings. First, our approach is the first to propose an integrated architecture capable of producing both PIs and well-calibrated value predictions in an end-to-end manner, without post-hoc optimization or increases to model complexity. Moreover, since our approach produces predictions for \textit{all} training set samples, \MethodName does not overfit to data values which were contained in their respective PIs. 

Secondly, \MethodName proposes a novel method for producing the value prediction. While previous studies either provided the middle of the PI \cite{qd, SQR} or the mean-variance \cite{deep_ensemble} as their value predictions, \MethodName's auxiliary head can produce any value within the PI, while also being distribution free (i.e., without making any assumptions regarding data distribution within the PI). By expressing the value prediction as a function of the upper and lower bounds, we ensure that the prediction \textit{always} falls within the PI without having to add additional penalty terms to the loss. Our use of a \textit{non-Gaussian observation model} enables us to produce predictions that are not in the middle of the interval and create representations that are more characteristic of many real-world cases, where the values within the PI is not necessarily uniformly distributed (see Section \ref{subsec:discussion_synthetic}).

\section{Method}


\subsection{System Architecture}
\begin{figure}[ht]
\centering
\includegraphics[width=0.8\textwidth]{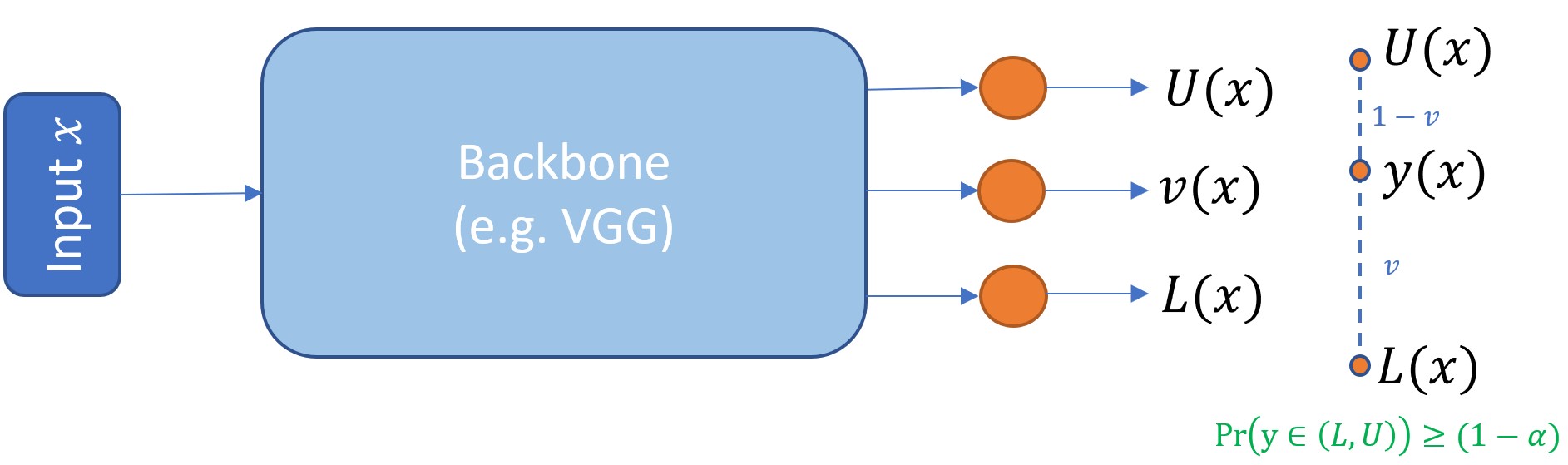}
\caption{The \MethodName schematic architecture}
\label{fig:method_archi} 
\end{figure}

The proposed architecture is presented in Figure \ref{fig:method_archi}. It consists of three components:

\noindent \textbf{Backbone block.} The main body block, consisting of a varying number of DNN layers or sub-blocks. The goal of this component is to transform the input into a latent representation that is then provided as input to the other components. It is important to note that \MethodName supports any architecture type (e.g., dense, convolutions) that can be applied to a regression problem. Moreover, pre-trained architectures can also be easily used, with \MethodName being added on top of the architecture for an additional short training. For example, we use pre-trained DenseNet architectures in our experiments.

\noindent \textbf{Upper \& lower-bound  heads.} \(L(x)\) and  \(U(x)\) produce the lower and upper bounds of the PI respectively, such that $\texttt{Pr}(L(x)\le y(x) \le U(x)) \ge 1-\alpha$ where \(y(x)\) is the value prediction and $1-\alpha$ is the predefined confidence level.

\noindent \textbf{Auxiliary head.} The auxiliary prediction head, \(v(x)\), enables us to produce a value prediction . \(v(x)\) does not produce the value prediction directly, but rather produces the \textit{relative weight that should be given to each of the two bounds}. We define the value prediction using,
    \begin{equation}
        y = v \cdot U + (1-v) \cdot L
    \end{equation}
where \(v \in (0,1)\). Aside from enabling us to directly produce a value prediction, the auxiliary head has additional advantages in terms of optimization and the prevention of overfitting. We elaborate further in Section \ref{subsec:auxHead}. 

\subsection{Network Optimization}
Our goal is to generate narrow PIs, measured by \textit{MPIW}, while maintaining the desired level of coverage \( PICP = 1-\alpha \). However, PIs that fail to capture their respective data point should not be encouraged to shrink further. Therefore, we define \textit{captured} \(MPIW\) (\(MPIW_{capt}\)) as the \(MPIW\) of only those points for which \( L_i \le y_{i} \le U_i \),
\begin{equation}\label{eq:mpiw_capt}
    MPIW_{capt} \defeq \frac{1}{c}\sum_{i=1}^{n} (U_i-L_i)\cdot k_i 
\end{equation}
where \(c = \sum_{i=1}^{n} k_i \).
Hence, we seek to minimize \(MPIW_{capt}\) subject to \(PICP \ge 1-\alpha\): 
\begin{equation} \label{eq:mpiw_capt_min_picp_max}
\begin{split}
 \theta^* = \argmin_{\theta} (MPIW_{capt, \theta}) 
 \ \ \text{s.t} \ \  PICP_{\theta} \ge 1-\alpha
\end{split}
\end{equation}
where \(\theta\) is the parameters of the neural net. To enforce the coverage constraint, we utilize a variant of the well-known Interior Point Method (IPM) \cite{IPM}, resulting in an unconstrained loss:  
\begin{equation} \label{eq:loss_qd}
\begin{split}
  \lossqd &= MPIW_{capt, \theta}+\sqrt{n} \cdot \lambda \Psi(1-\alpha-PICP_{\theta})\\ 
  \Psi(x) & \defeq \max(0,x)^2
\end{split}
\end{equation}
where \(\Psi\) is a quadratic penalty function, \(n\) is the batch size which was included because a larger sample size would increase confidence in the value of PICP (thus increasing the loss) and \(\lambda\) is a hyperparameter controlling the relative importance of width vs. coverage. We use a default value of \(\lambda = 15\) in all our experiments, and perform further analysis of this parameter in Section \ref{subsec:hyperparam_influence}. 

In practice, optimizing the loss with a discrete version of \(\textbf{k}\) (see eq.~\ref{eq:mpiw_capt}) fails to converge, because the gradient is always positive for all possible values. We therefore define a continuous version of \(\textbf{k}\), denoted as
\(
\textbf{k$_{soft}$} = \sigma(s \cdot (\textbf{y} - \textbf{L})) \odot \sigma(s\cdot (\textbf{U}-\textbf{y})) 
\),
where \(\sigma\) is the sigmoid function, and \(s>0\) is a softening factor. The final version of \(\lossqd\) uses the continuous and discrete versions of \(\textbf{k}\) in its calculations of the \(PICP\) and \(MPIW_{capt}\) metrics, respectively. The discrete \(\textbf{k}\) enables us to assign a score of zero to points outside the interval, while \(\textbf{k$_{soft}$}\) produces continuous values that enable gradient calculations. 

\subsection{The Auxiliary Head}
\label{subsec:auxHead}
The auxiliary head \(v(x)\) serves three goals: first, it enables \MethodName to select any point within the PI as the value prediction, an approach that produces superior results to choosing the middle of the PI, as done in previous works. Secondly, since we produce a value prediction for every point, and not only to those that fall within their PI, we enable \MethodName to learn from the entire dataset. By doing so we avoid the overfitting problem described in Section \ref{sec:problemFormulation}. Thirdly, the value prediction is expressed as a function of the upper and lower bounds, thus ensuring that all three outputs -- upper, lower, and auxiliary -- are optimized jointly. This setting also ensures the integrity of our output, since the value prediction \textit{always} falls within the interval. To optimize the output of \(v(x)\), we minimize:
\begin{equation}\label{eq:h_loss}
    \mathcal{L}_v = \frac{1}{n}\sum_{i=1}^{n} \ell(v_i \cdot U_i + (1-v_i)\cdot L_i, \ y_i) 
\end{equation}
where \(\ell\) can be any regression objective against the ground-truth. Our final loss function is a convex combination of \(\lossqd\), and the auxiliary loss $\mathcal{L}_v$. Thus, the overall training objective is:
\begin{equation}
    \label{eq:LossPiven}
    \mathcal{L}_{\MethodName} = \beta \lossqd + (1-\beta) \mathcal{L}_v 
\end{equation}
where \(\beta\) is a hyperparameter that balances the two goals of our approach: producing narrow PIs and accurate value predictions. In all our experiments, we chose to assign equal priorities to both goals by setting \(\beta = 0.5\). An analysis of the effects of various values of \(\beta\) is presented in Section \ref{subsec:hyperparam_influence}.

\subsection{Using Ensembles to Estimate Model Uncertainty}
Ensembles are commonly used to reduce uncertainty and improve performance. In our setting, each ensemble model produces a PI and a value prediction. These outputs need to be combined into a single PI and value prediction , thus capturing both the aleatoric and parametric uncertainties.

While PIs produced by a single \MethodName architecture could be used to capture aleatoric uncertainty, an ensemble of \MethodName architectures can capture the uncertainty of the PI itself. We consider the aggregation proposed by \cite{deep_ensemble}, as it is used by SOTA studies in the field:

Given an ensemble of \(m\) NNs trained with $\mathcal{L}_{PIVEN}$, let $\tilde{U}$, $\tilde{L}$ represent the PI, and $\tilde{v}$, $\tilde{y}$  represent the ensemble’s auxiliary and value prediction. We calculate the PI uncertainty and use the ensemble to generate the PIs and 
value predictions as follows: 
\hspace*{-4em}
\begin{tabular}{p{7cm}p{7cm}}
{\makeatletter\CT@everycr{\the\everycr}{\begin{align}
&\bar{U_i} = \frac{1}{m}\sum_{j=1}^m U_{ij} \\
&\tilde{{U_{i}}} = \bar{{U_{i}}} + z_{\alpha \slash 2} \cdot \sigma_{U_{i}}
\end{align}}}
&
{\makeatletter\CT@everycr{\the\everycr}{\begin{align}
&\sigma_{PI}^2 = \sigma_{U_{i}}^2 = \frac{1}{m-1}\sum_{j=1}^m(U_{ij}-\bar{U_{i}})^2 \\ 
&\tilde{y_i} = \frac{1}{m}\sum_{j=1}^m v_{ij} \cdot U_{ij} + (1-v_{ij})\cdot L_{ij}
\end{align}}}
\end{tabular}


where (\(U_{ij}\) \(L_{ij}\)) and \(v_{ij}\) are the PI and the auxiliary prediction for data point \(i\), for NN \(j\). A similar procedure is done for \(\tilde{L_{i}}\), subtracting \(z_{\alpha \slash 2} \cdot \sigma_{L_{i}}\), where \(z_{\alpha \slash 2}\) is the \textit{Z} score for confidence level \(1-\alpha\).

\section{Evaluation}
\label{subsec:evaluation}

\subsection{Baselines}
\label{subsec:baselines}

We compare our performance to three top-performing baselines from recent years: 

\noindent \textbf{Quality Driven PI (QD) \cite{qd}.} QD produces PIs that minimize a smooth combination of the PICP/MPIW metrics without considering the value prediction task in its objective function. Its reported results make it state-of-the-art in terms of PI width and coverage.

\noindent \textbf{Quality Driven Plus (QD+) \cite{salem}}. A recently proposed enhancement to QD, consisting of two improvements: \textit{a)} additional terms to the loss functions, used to produce value predictions, and; \textit{b)} a previously-proposed post-optimization aggregation method called SNM \cite{split_normal}. While effective, SNM has a very high computational overhead (as stated by the original authors \cite{split_normal}), which makes QD+ impractical for large datasets such as our large image datasets, discussed in Section \ref{subsec:datasets}. We therefore run QD+ using the same optimization method as QD and \MethodName. This setting provides a level playing field and enables us to compare the performance of each method's loss function, which is the main contribution of these respective studies. Nonetheless, for the sake of completeness, we also present all the reported results of the original QD+ algorithm together with our own experiments. These results, on the UCI datasets, are presented in Section \ref{subsec:evalUCI}.

\noindent \textbf{Deep Ensembles (DE) \cite{deep_ensemble}.} This work combines individual conditional Gaussian distribution with adversarial training, and uses the models' variance to compute prediction intervals. Because DE outputs distributions instead of PIs, we first convert it to PIs, and then compute PICP and MPIW. DE's reported results make one of the top performers with respect to value prediction.


\subsection{The Evaluated Datasets}
\label{subsec:datasets}


\textbf{Synthetic datasets.} We generate two datasets with skewed (i.e., neither Gaussian nor uniform) value distributions. We use these datasets to demonstrate \MethodName's effectiveness compared to methods that only return the middle of the interval as a value prediction (namely, QD). Our datasets are the Sine and skewed-normal distributions.

\textbf{IMDB age estimation dataset.} The IMDB-WIKI dataset \cite{imdb_ds} is currently the largest age-labeled facial dataset available. Our dataset consists of 460,723 images from 20,284 celebrities, and the regression goal is to predict the age of the person in the image. This dataset is known to contain noise (i.e., aleatoric uncertainty), thus making it relevant (despite usually being used for pre-training).

\textbf{RSNA pediatric bone age dataset.} This dataset is a popular medical imaging dataset consisting of X-ray images of children's hands \cite{rsna}. The regression task is predicting one's age from one's bone image. The dataset contains 12,611 training images and 200 test set images. 

\textbf{UCI Datasets.} A set of 10 datasets used by several state-of-the-art studies \cite{mc_dropout, pbp, deep_ensemble, qd}. Commonly used as benchmark for new studies.

Our selection of datasets is designed to achieve two goals: \textit{a)} demonstrate that \MethodName is applicable to multiple and more challenging domains (e.g., images), unlike previous studies that were only applied to tabular datasets and; \textit{b)} show that \MethodName significantly outperforms PI-based methods (5\%--17\% improvement in MPIW), while maintaining comparable performance to non-PI methods.

\subsection{Experimental Setup}
\label{subsec:expSetup}

We evaluate all baselines using their reported deep architectures and hyperparameters. For full experimental details, please see Appendix \ref{sec:appendix_exp_setup}. We ran our experiments using a GPU server with two NVIDIA Tesla P100. Our code is implemented using TensorFlow \cite{tf}, and is available online\footnote{\url{https://github.com/anonymous/anonymous}}.

\textbf{Synthetic datasets.} We used a neural net with one hidden layer of 100 nodes with ReLU \cite{relu}. All methods trained until convergence using the same initialization, loss and default hyperparameters: \(\alpha=0.05, \beta = 0.5, \lambda = 15.0, s = 160.0\), with random seed set to 1. The generated data consisted of 100 points sampled from the interval \([-2, 2]\). We used \(\alpha = 100\) as the skewness parameter.

\textbf{IMDB age estimation dataset.} We use the DenseNet architecture \cite{densenet} as the backbone block, then add two dense layers. We apply the data preprocessing used by \cite{SSRNet}. We use 5-fold cross validation. 

\textbf{RSNA bone age dataset.} We use VGG-16 \cite{vgg} as the backbone block, with weights pre-trained on ImageNet. We add two convolutional layers followed by a dense layer, and perform additional training. This dataset has 200 predefined test images.

\textbf{UCI datasets.} We use the same experimental setup used by our baselines, as proposed by \cite{pbp}. Results are averaged on 20 random 90\%/10\% splits of the data, except for the ``Year Prediction MSD" and ``Protein'', which were split once and five times respectively. Our network architecture is identical to that of our two baselines: one hidden layer with ReLU activation function, the Adam optimizer \cite{adam}, and ensemble size $M=5$. Input and target variables are normalized to zero mean and unit variance, like in previous works.

\subsection{Synthetic Datasets: Value Prediction in Skewed Distributions}
\label{subsec:discussion_synthetic}

\begin{figure}[ht]
\centering
\includegraphics[width=0.8\textwidth]{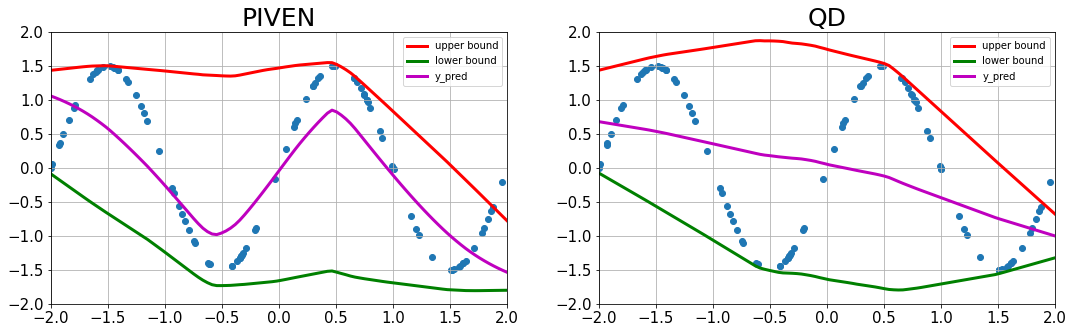}
\caption{Comparison of value prediction for QD vs. \MethodName method, where \(\bm{y}\) was generated by \(f(x) = 1.5 \sin(x)\).}
\label{fig:synthetic_sin} 
\end{figure}

The goal of this evaluation is to demonstrate the shortcomings of returning the middle of the interval as value prediction, as done by methods such QD. 
As noted in \cite{qd}, QD ``breaks the distribution-free assumption'', because such a selection is only optimal for Gaussian or uniformly based distributions. To test \MethodName's ability to outperform QD in skewed distributions, we evaluate both approaches on the Sine and skewed-normal distributions. The results, presented in Figure \ref{fig:synthetic_sin} (for Sine) and in Figure \ref{fig:appendix_synthetic_skew_normal} in the appendix (for skewed-normal), clearly illustrate \MethodName's superior ability to adapt to multiple value distributions, through the use of its auxiliary head.


\subsection{Large-Scale Image Datasets}


Both bone age and IMDB are large and noisy image datasets. We follow \cite{rio} and add an additional baseline -- denoted as NN -- which is a single dense layer on top of existing architectures (DenseNet/VGG). This layer outputs value prediction using the MSE metric. Instead of RMSE, we use \textit{mean absolute error} (MAE), which was the datasets' chosen metric. For IMDB, it is clear that \MethodName outperforms the other PI-driven methods (QD and QD+) by wide margins---20\% in MPIW and 43\% in MAE---while also outperforming NN. DE performed best for this dataset, with a small margin over \MethodName, but this likely due to the fact that age distribution is a Gaussian, and therefore fully compatible with this baseline's assumptions. \MethodName obtains comparable results to DE without relying on any distributional assumptions. It is also important to note that \MethodName performs significantly better when encountering outliers -- very young or very old persons -- as shown in Figure \ref{fig:imdb_outliers} in the appendix. In bone age, \MethodName outperforms all baselines in terms of MAE. Our approach fares slightly worse compared to QD on the MPIW metric, but that is likely due to its higher coverage (i.e., PICP), which means it had to contend with more samples, particularly those which were difficult to classify (the same applies for QD+, which likely contributed to its higher error rates). 

\begin{table}[ht]
    \footnotesize
    \RawFloats
    \centering
    \caption{Results on the RSNA bone age and IMDB age estimation datasets.}
    \resizebox{0.55\textwidth}{!}{
     \begin{tabular}{c|c|c|c|c} 
    \toprule
     Datasets & Method & PICP & MPIW & MAE \\ 
     \midrule
      \multirow{5}{*}{IMDB age} & NN & NA & NA & 7.08 $\pm$ 0.03\\
                               & QD & 0.92 $\pm$ 0.01 & 3.47 $\pm$ 0.03 & 10.23 $\pm$ 0.12 \\
                               & QD+ &\textbf{ 1.00 $\pm$ 0.00 }& 5.01 $\pm$ 0.11 & 10.08 $\pm$ 0.03 \\
                               & DE & \textbf{0.95 $\pm$ 0.01} & \textbf{2.61 $\pm$ 0.05} & \textbf{6.66 $\pm$ 0.06} \\
                               & \MethodName & \textbf{0.95 $\pm$ 0.01} & 2.87 $\pm$ 0.04 & 7.03 $\pm$ 0.04 \\
                               
    \midrule                    
    \multirow{5}{*}{Bone age} & NN & NA & NA & 18.68\\
                                & QD & 0.90 & \textbf{1.99} & 20.24 \\
                                & QD+ &\textbf{ 1.00} & 5.24 & 25.18   \\
                                & DE & \textbf{0.93} & 2.17 & 18.69 \\
                                & \MethodName & \textbf{0.93}  & 2.09 & \textbf{18.13} \\
    
      \bottomrule
    \end{tabular}
    }
    \label{tab:large_ds_resultsS}
\end{table}

\begin{table}[ht]
\RawFloats
\caption{Regression results for benchmark UCI datasets comparing MPIW, and RMSE. The grey column contains the results for QD+ with their proposed aggregation method, as reported by the authors.}
\label{table:uci_table}
\centering
\resizebox{\textwidth}{!}{
    \begin{tabular}{l|cccca|cccca}
    \toprule
    \multirow{2}{*}{} &
      \multicolumn{5}{c|}{\textbf{MPIW}} &
      \multicolumn{5}{c}{\textbf{RMSE}} 
      \\
      Datasets & {DE} & {QD} & {QD+} & {\MethodName} & {QD+(Reported)} & {DE} & {QD} & {QD+} & {\MethodName} & {QD+(Reported)} \\
      \midrule
    Boston & \textbf{0.87 $\pm$ 0.03} & 1.15 $\pm$ 0.02 & 4.98 $\pm$ 0.81 & 1.09 $\pm$ 0.01 & 1.58 $\pm$ 0.06 & \textbf{2.87 $\pm$ 0.19} & 3.39 $\pm$ 0.26 & 6.12 $\pm$ 0.73 & 3.13 $\pm$ 0.21 & 0.12 $\pm$ 0.01 \\
    Concrete & \textbf{1.01 $\pm$ 0.02} & 1.08 $\pm$ 0.01 & 3.43 $\pm$ 0.26 & 1.02 $\pm$ 0.01 & 0.99 $\pm$ 0.04 &\textbf{ 5.21 $\pm$ 0.09 }& 5.88 $\pm$ 0.10 & 11.87 $\pm$ 0.83 & 5.43 $\pm$ 0.13 & 0.05 $\pm$ 0.00 \\
    Energy & 0.49 $\pm$ 0.01 & 0.45 $\pm$ 0.01 & 1.52 $\pm$ 0.18  &  \textbf{0.42 $\pm$ 0.01} & 0.29 $\pm$ 0.01 & 1.68 $\pm$ 0.06 & 2.28 $\pm$ 0.04 & 7.05 $\pm$ 0.56  & \textbf{ 1.65 $\pm$ 0.03 } & 0.00 $\pm$ 0.00\\
    Kin8nm & 1.14 $\pm$ 0.01 & 1.18 $\pm$ 0.00 & 2.64 $\pm$ 0.19  & \textbf{1.10 $\pm$ 0.00 } & 1.07 $\pm$ 0.01 & 0.08 $\pm$ 0.00 & 0.08 $\pm$ 0.00 & 0.19 $\pm$ 0.01  & \textbf{0.07 $\pm$ 0.00} & 0.06 $\pm$ 0.00\\
    Naval & 0.31 $\pm$ 0.01 & 0.27 $\pm$ 0.00 & 2.85 $\pm$ 0.15  & \textbf{0.24 $\pm$ 0.00 } & 0.09 $\pm$ 0.00 & \textbf{0.00 $\pm$ 0.00} & \textbf{0.00 $\pm$ 0.00} & 0.01 $\pm$ 0.00 & \textbf{0.00 $\pm$ 0.00} &  0.00 $\pm$ 0.00 \\
    Power & 0.91 $\pm$ 0.00 & \textbf{0.86 $\pm$ 0.00} & 1.51 $\pm$ 0.06  &  \textbf{0.86 $\pm$ 0.00 } & 0.80 $\pm$ 0.00 & \textbf{3.99 $\pm$ 0.04} & 4.14 $\pm$ 0.04 & 12.00 $\pm$ 0.61  &  4.08 $\pm$ 0.04 &  0.05 $\pm$ 0.00\\
    Protein & 2.68 $\pm$ 0.01 & \textbf{2.27 $\pm$ 0.01} & 3.30 $\pm$ 0.03 & \textbf{2.26 $\pm$ 0.01}  & 2.12 $\pm$ 0.01 & \textbf{4.36 $\pm$ 0.02} & 4.99 $\pm$ 0.02 &  5.10 $\pm$ 0.09 &  \textbf{4.35 $\pm$ 0.02} & 0.36 $\pm$ 0.00 \\
    Wine & 2.50 $\pm$ 0.02 & \textbf{2.24 $\pm$ 0.02} & 4.89 $\pm$ 0.12 & \textbf{2.22 $\pm$ 0.01  } & 2.62 $\pm$ 0.06 & \textbf{0.62 $\pm$ 0.01} & 0.67 $\pm$ 0.01 & 0.70 $\pm$ 0.04  &  \textbf{0.63 $\pm$ 0.01} &  0.62 $\pm$ 0.02 \\
    Yacht & 0.33 $\pm$ 0.02 & 0.18 $\pm$ 0.00 & 1.25 $\pm$ 0.23  & \textbf{0.17 $\pm$ 0.00 } & 0.12 $\pm$ 0.00 & 1.38 $\pm$ 0.07 & 1.10 $\pm$ 0.06 & 8.19 $\pm$ 1.26 &  \textbf{0.98 $\pm$ 0.07}  &  0.00 $\pm$ 0.00 \\
    MSD & 2.91 $\pm$ NA & 2.45 $\pm$ NA & 4.25 $\pm$ NA  & \textbf{2.42 $\pm$ NA} & 2.34 $\pm$ NA & 8.95 $\pm$ NA & 9.30 $\pm$ NA & 10.11 $\pm$ 0.00  & \textbf{8.93 $\pm$ NA} &  0.64 $\pm$ NA \\
    \bottomrule
  \end{tabular}
  }
\end{table}

\subsection{UCI Datasets Results}
\label{subsec:evalUCI}

We evaluate the UCI datasets with \(\alpha = 0.05\), as done in previous works, with additional experiments later in this section. Since all methods achieved the required PICP metric, we present these results in Table \ref{table:appendix_uci_table_pis} in the appendix. The results for the MPIW and RMSE metrics are presented in Table \ref{table:uci_table}.
In terms of PI-quality, \MethodName achieves top performance eight of of ten datasets (QD achieves comparable results in three), while achieving comparable performance to DE in one of the remaining datasets. For the RMSE metric, it is clear that \MethodName and DE are the top performers, with the former achieving the best results in five datasets, and the latter in four. These results are notable because DE does not need to simultaneously optimize for PI and value prediction, while \MethodName aims to balance these two goals. The QD and QD+ baselines trail behind the other methods in all datasets but one. QD's performance is not surprising given that the focus of this approach is PI generation rather than value prediction, and QD+'s need to optimize a larger number of terms in the loss function. Finally, we also perform additional experiments for DE with ensemble size $M=1$, with the results presented in the appendix (Table \ref{table:de_one_uci_table}).

For completeness, we also include the results of QD+ with their proposed aggregation method, as reported in \cite{salem}. These results appear as QD+(reported) in Table \ref{table:uci_table}. The large difference in performance between the two QD+ variants leads us to conclude that QD+'s multiple terms in the loss function increase model complexity, which in turn makes their computationally-heavy optimization necessary. Using this optimization, however, makes QD+ impractical for large datasets. This conclusion is supported by the authors of \cite{salem}: ``A potential drawback of SNM is the apparent computational overhead of fitting the split normal mixture when the number of samples is large''. 

\textbf{Performance in varying levels of coverage.} In Section \ref{subsec:motivation} we hypothesize that \MethodName's auxiliary head makes it less prone to overfitting by enabling it to train on the entire training set rather than only on the data points captured within their PIs, as done by current SOTA PI-based approaches such as QD. To test this hypothesis, we evaluate \MethodName and QD on varying levels of coverage, i.e., where we require different percentages of the dataset to be captured by their PIs. Summary results for MPIW are presented in Table \ref{table:alpha_mpiw_improve_pct} and full results are in Appendix \ref{sec:appendix_alpha_full_results}. It is clear that as coverage decreases, \MethodName's relative performance to QD improves. While \MethodName's advantage is more pronounced in lower coverage rates, it persists in all coverage levels. Next, we compare the two methods using the RMSE metric. The results of our analysis -- an example of which is presented in Figure \ref{fig:yacht_msd_rmse_alpha} and the full results in Appendix \ref{sec:appendix_alpha_full_results} -- clearly show that while QD's performance deteriorates along with the coverage levels, \MethodName's remains largely unchanged.

\begin{table*}[ht]
    \RawFloats
	\begin{minipage}{0.5\linewidth}
		\centering
		\resizebox{\textwidth}{!}{
		\begin{tabular}{l|cccccc}
        \toprule
        \multirow{2}{*}{} &
        \multicolumn{6}{c}{alpha} \\ 
          Datasets & 0.05 & 0.10 & 0.15 & 0.20 & 0.25 & 0.30  \\
          \midrule
            Boston & 7\% & 7\% & 6\% & 7\% & 8\% & 8\% \\
            Concrete & 7\% & 7\% & 8\% & 7\% & 12\% & 13\% \\
            Energy & 6\% & 12\% & 8\% & 17\% & 19\% & 27\% \\
            Kin8nm & 5\% & 7\% & 9\% & 19\% & 13\% & 15\% \\
            Naval & 10\% & 16\% & 16\% & 14\% & 17\% & 23\% \\
            Power & 0\% & 0\% & 1\% & 1\% & 1\% & 1\% \\
            Protein & 1\% & 1\% & 2\% & 3\% & 2\% & 2\% \\
            Wine & 0\% & 1\% & 2\% & 2\% & 4\% & 4\% \\
            Yacht & 8\% & 12\% & 14\% & 20\% & 20\% & 21\% \\
            MSD & 1\% & -3\% & -3\% & 1\% & -5\% & -2\% \\
            \midrule
            \midrule
            \textbf{Average} & \textbf{4.5}\% & \textbf{6.0}\% & \textbf{6.3}\% & \textbf{9.1}\% & \textbf{9.1}\% & \textbf{11.2}\% \\
        \bottomrule
        \end{tabular}
        }
    \vspace{3mm}
    \caption{MPIW improvement in percentages (\MethodName relative to QD) as the coverage decreases (i.e, alpha increases).} 
    \label{table:alpha_mpiw_improve_pct}
	\end{minipage}  \hspace{2mm}
	\begin{minipage}{0.5\linewidth}
		\centering
		\includegraphics[width=0.7\textwidth]{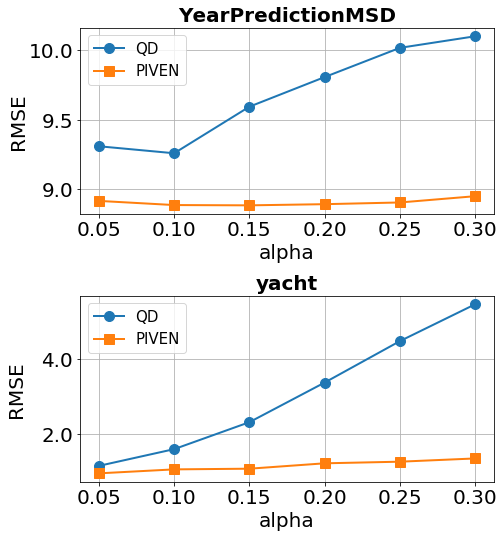}
		\vspace{3mm}
		\captionof{figure}{RMSE as function of \(\alpha\) on two of the UCI datasets. See Appendix \ref{sec:appendix_alpha_full_results} for full results.}
		\label{fig:yacht_msd_rmse_alpha}
	\end{minipage}
\end{table*}

\section{Analysis}
\label{sec:discussion}

\subsection{Auxiliary Head Analysis}
\label{subsec:auxiliary_head_analysis}
In Section \ref{subsec:motivation} we describe our rationale in expressing the value prediction as a function of the upper and lower bounds of the interval. To further explore the benefits of our auxiliary head \(v\), we evaluate two variants of \MethodName. In the first variant, denoted as \OnlyRmse (point-only optimization), we decouple the value prediction from the PI. The loss function of this variant is $\lossqd + \ell(v,y_{true})$ where $\ell$ is set to be MSE loss. In doing so, we compare PIVEN to an approach where the value prediction is not expressed as a function of the bounds (as was done in QD+ \cite{salem}). In the second variant, denoted \Vhalf (middle of interval), the value prediction produced by the model is always the middle of the PI (in other words, $v$ is set to 0.5). While both return the middle of the PI as value prediction, MOI differs from QD by an additional component in its loss function.

The results of our analysis are presented in Table \ref{table:ablation_uci_table_pis_rmse}, which contains the results of the MPIW and RMSE metrics (PICP values are identical, see Appendix \ref{sec:appendix_ablation}). The full \MethodName significantly outperforms the two other variants. We therefore conclude that both novel aspects of our approach---the simultaneous optimization of PI-width and RMSE, and the ability to select any value on the PI as the value prediction---contribute to \MethodName's performance. Finally, note that while inferior to \MethodName, both \OnlyRmse and \Vhalf outperform the QD baseline in terms of MPIW, while being equal or better for RMSE. 

\begin{table}[ht]
\RawFloats
\caption{Analysis comparing MPIW and RMSE. Results were analyzed as in Table \ref{table:uci_table}.}
\label{table:ablation_uci_table_pis_rmse}
\centering
\resizebox{0.7\textwidth}{!}{
  \begin{tabular}{l|ccc|ccc}
    \toprule
    \multirow{2}{*}{} &
      \multicolumn{3}{c|}{MPIW} &
      \multicolumn{3}{c}{RMSE} 
      \\
      Datasets & {\OnlyRmse } & {\Vhalf} & {\MethodName} & {\OnlyRmse} & {\Vhalf} & {\MethodName} \\
      \midrule
    Boston & \textbf{1.09 $\pm$ 0.02} & 1.15 $\pm$ 0.02 & \textbf{1.09 $\pm$ 0.01} & 3.21 $\pm$ 0.24 & 3.39 $\pm$ 0.27 & \textbf{3.13 $\pm$ 0.21} \\
    Concrete & \textbf{1.02 $\pm$ 0.01} & 1.07 $\pm$ 0.01 & \textbf{1.02 $\pm$ 0.01} & 5.55 $\pm$ 0.11 & 5.73 $\pm$ 0.10 & \textbf{5.43 $\pm$ 0.13} \\
    Energy & \textbf{0.42 $\pm$ 0.01} & 0.45 $\pm$ 0.01 & \textbf{0.42 $\pm$ 0.01} & 2.16 $\pm$ 0.04 & 2.27 $\pm$ 0.04 & \textbf{1.65 $\pm$ 0.03} \\
    Kin8nm & 1.13 $\pm$ 0.00 & 1.17 $\pm$ 0.00 & \textbf{1.10 $\pm$ 0.00} & 0.08 $\pm$ 0.00 & 0.08 $\pm$ 0.00 & \textbf{0.07 $\pm$ 0.00 } \\
    Naval & \textbf{0.24 $\pm$ 0.00} & 0.30 $\pm$ 0.02 & \textbf{0.24 $\pm$ 0.00} & \textbf{0.00 $\pm$ 0.00} & \textbf{0.00 $\pm$ 0.00} & \textbf{0.00 $\pm$ 0.00} \\
    Power &\textbf{ 0.86 $\pm$ 0.00} & \textbf{0.86 $\pm$ 0.00} & \textbf{0.86 $\pm$ 0.00} & 4.13 $\pm$ 0.04 & 4.15 $\pm$ 0.04 & \textbf{4.08 $\pm$ 0.04} \\
    Protein & \textbf{2.25 $\pm$ 0.01} & 2.27 $\pm$ 0.01 & \textbf{2.26 $\pm$ 0.01} & 4.78 $\pm$ 0.02 & 4.99 $\pm$ 0.01 & \textbf{4.35 $\pm$ 0.02} \\
    Wine & 2.24 $\pm$ 0.01 & 2.23 $\pm$ 0.01 & \textbf{2.22 $\pm$ 0.01} & 0.64 $\pm$ 0.01 & 0.67 $\pm$ 0.01 & \textbf{0.63 $\pm$ 0.01} \\
    Yacht & 0.18 $\pm$ 0.00 & 0.19 $\pm$ 0.01 & \textbf{0.17 $\pm$ 0.00}  & 0.99 $\pm$ 0.07 & 1.15 $\pm$ 0.08 & \textbf{0.98 $\pm$ 0.07} \\
    MSD & \textbf{2.42 $\pm$ NA} & 2.43 $\pm$ NA & \textbf{2.42 $\pm$ NA} & 9.10 $\pm$ NA & 9.25 $\pm$ NA &\textbf{ 8.93 $\pm$ NA} \\
    \bottomrule
  \end{tabular}
  }
\end{table}

\subsection{Hyperparameters Analysis}
\label{subsec:hyperparam_influence}

\begin{figure}[ht]
\centering
\includegraphics[width=\textwidth]{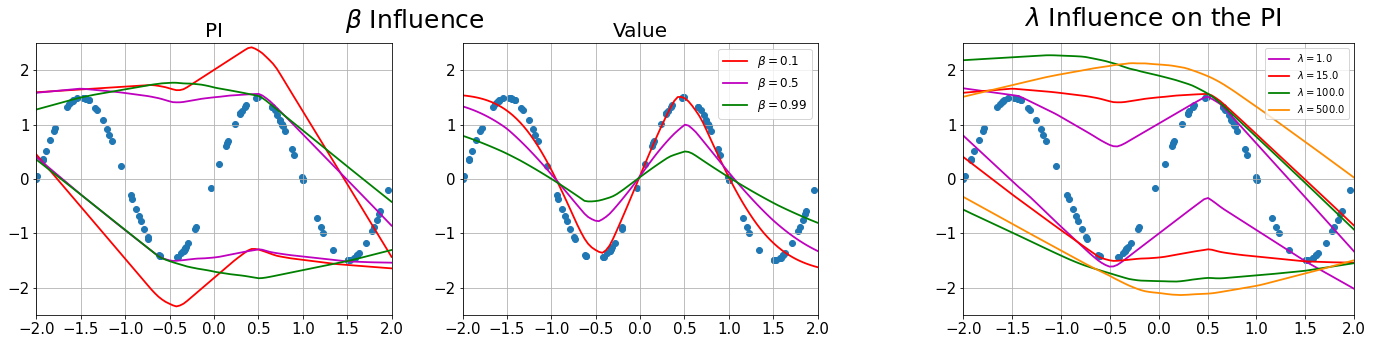}
\caption{(Left) An analysis of the effect of varying \(\beta\) values on \MethodName's performance in terms of PI width and value prediction. (Right) The effect of various \(\lambda\) values on the tightness of the PI}
\label{fig:hyperparam_influence}
\end{figure}

Two parameters govern the behavior of our approach: $\beta$  (Equation  \ref{eq:LossPiven}) and \(\lambda\) (Equation \ref{eq:loss_qd}). We use the Sine function, whose value fluctuations are challenging to any PI-producing approach, and will therefore make identifying relevant patterns easier. $\beta$ balances the two goals of our approach: narrow PIs and accurate value predictions. As shown in Equation \ref{eq:LossPiven}, $\beta$ determines the weight assigned to each goal in \MethodName's loss function. Therefore, for large values of $\beta$ we expect \MethodName to put greater emphasis on optimizing the PI at the expense of the value prediction. For small values of $\beta$, we expect the opposite. Figure \ref{fig:hyperparam_influence} (Left) confirms our expectations. It is clear that $\beta=0.99$ produces the most accurate PIs, but its value predictions is the least accurate. The opposite applies for $\beta=0.1$
This analysis also shows that $\beta=0.5$ strikes a better balance, as it optimizes for both tasks simultaneously.

The goal of \(\lambda\) is to balance capturing as many samples as possible within our PIs (PICP) with the conflicting goal of producing tight PIs (MPIW). We expect that for small \(\lambda\) values, \MethodName will attempt to produce tight PIs with less emphasis on coverage. For large \(\lambda\) values, we expect the opposite. Results in Figure \ref{fig:hyperparam_influence} (Right) confirm our expectations.

\newpage
\section{Conclusions}
\MethodName is a novel architecture for combining the generation of prediction intervals together with specific value predictions in an end-to-end manner. \MethodName optimizes these two goals simultaneously and is able to achieve improved results in both. Our performance is enhanced by two factors: first, its value-prediction and PI-prediction are connected and optimized simultaneously. Secondly, \MethodName does not ignore but rather trains on data points that are not captured by the PI. 

\small

\bibliographystyle{plain}

\bibliography{refs}

\clearpage
\appendix
\section{Experimental Setup}
\label{sec:appendix_exp_setup}
In this section we provide full details of our dataset preprocessing and experiments presented in the main study. Our code is available online \footnote{\url{https://github.com/anonymous/anonymous}}

\subsection{Synthetic Datasets}
\label{sec:appendix_synthetic}
\begin{figure}[ht]
\centering
\includegraphics[width=0.8\textwidth]{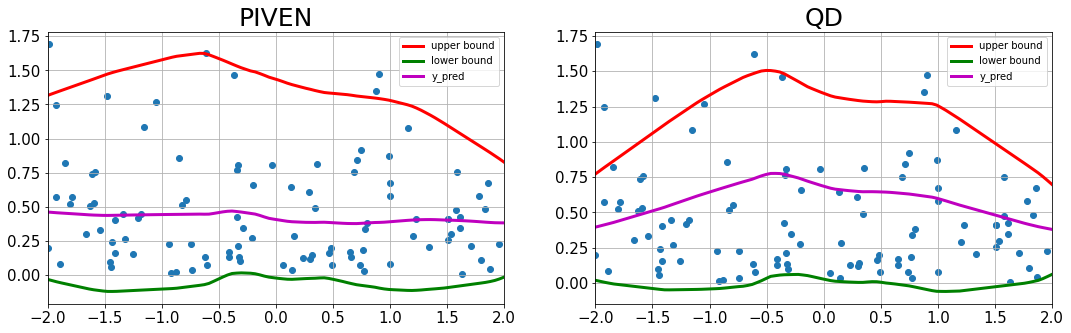}
\caption{Value prediction comparison, where \(\bm{y}\) is sampled from skewed-normal distribution \(f(x; \alpha) = 2 \phi (x)\Phi(\alpha \cdot x )\) where \(\phi (x)\) and \(\Phi(\alpha \cdot x )\) denote the \(\mathcal{N}(0,1)\) density and distribution function respectively; the parameter  \(\alpha\) denotes the skewness parameter.}
\label{fig:appendix_synthetic_skew_normal} 
\end{figure}

For the qualitative training method comparison, QD vs.
\MethodName (Section \ref{subsec:discussion_synthetic}), all NNs used ReLU activations and 100 nodes in one hidden layer. Both methods trained until convergence using the same initialization, loss and default hyperparameters, as described in \cite{qd}: \(\alpha=0.05, \beta = 0.5, \lambda = 15.0, s = 160.0\), and the random seed was set to 1.
The generated data consisted of 100 points sampled uniformly from the interval \([-2, 2]\). We used \(\alpha = 100\) as the skewness parameter. In Figures \ref{fig:synthetic_sin} and \ref{fig:appendix_synthetic_skew_normal} we present a comparison on the Sine and skewed-normal distributions.

\subsection{Dataset Preprocessing and Full Experimental Setup}
In addition to the ten benchmark datasets used by all recent studies in the field, we evaluated \MethodName on two large image datasets. Due to the size of the datasets and the nature of the domain, preprocessing was required. We provide the full details of the process below.

\textbf{UCI datasets.} \ \ 
For the UCI datasets, we used the experimental setup proposed by \cite{pbp}, which was also used in all the two baselines described in this study. All datasets were averaged on 20 random splits of the data, except for the ``Year Prediction MSD" and ``protein" datasets. Since ``Year Prediction MSD" has  predefined fixed splits by the provider, only one run was conducted. For "protein", 5 splits were used, as was done in previous work. Number of samples and features for each dataset is presented in Table \ref{table:uci_samples}. We used identical network architectures to those described in previous works: one dense layer with ReLU \cite{relu}, containing 50 neurons for each network. In the ``Year Prediction MSD" and ``protein" datasets where NNs had 100 neurons. Regarding train/test split and hyperparameters, we employ the same setup as \cite{qd}: train/test folds were randomly split 90\%/10\%, input and target variables were normalized to zero mean and unit variance.  The softening factor was constant for all datasets, $s = 160.0$. For the majority of the datasets we used $\lambda = 15.0$, except for ``naval", ``protein", ``wine" and ``yacht" where $\lambda$ was set to 4.0, 40.0, 30.0 and 3.0 respectively.
The value of the parameter $\beta$ was set to 0.5. The Adam optimizer \cite{adam} was used with exponential decay, where learning rate and decay rate were tuned. Batch size of 100 was used for all the datasets, except for ``Year Prediction MSD" where batch size was set to 1000. Five neural nets were used in each ensemble, using parameter re-sampling. The objective used to optimized \(v\) was Mean Square Error (MSE) for all datasets.  We also tune $\lambda$, initializing variance, and number of training epochs using early stopping. To ensure that our comparison with the state-of-the-art baselines is accurate, we first set the parameters of our neural nets so that they produce the results reported in \cite{qd}. We then use the same parameter configurations in our experiments of \MethodName.

\begin{table}[ht]
\RawFloats
\caption{Number of samples and features for UCI datasets}
\label{table:uci_samples}
\centering
\resizebox{0.3\textwidth}{!}{
  \begin{tabular}{l|c|c}
    \toprule
      Dataset & \# samples & \# features \\
      \midrule
    Boston & 506 & 14 \\
    Concrete & 1030 & 9 \\
    Energy & 768 & 9 \\
    Kin8nm & 8192 & 9 \\
    Naval & 11934 & 18 \\
    Power & 9568 & 5 \\
    Protein & 45730 & 10 \\
    Wine & 1599 & 12 \\
    Yacht & 308 & 7 \\
    MSD & 515345 & 90 \\
    \bottomrule
  \end{tabular}
  }
\end{table}

\textbf{IMDB age estimation dataset}\footnote{\url{https://data.vision.ee.ethz.ch/cvl/rrothe/imdb-wiki/}}.\ \ 
For the IMDB dataset, we used the DenseNet architecture \cite{densenet} as a feature extractor. On top of this architecture we added two dense layers with dropout. The sizes of the two dense layers were 128 and 32 neurons respectively, with a dropout factor of 0.2, and ReLU activation \cite{relu}. In the last layer, the biases of the PIs were initially set to  $[5.0, -5.0]$ for the upper and lower bounds respectively. We used the data preprocessing similar to that of previous work \cite{SSRNet, zhang}: all face images were aligned using facial landmarks such as eyes and the nose. After alignment, the face region of each image was cropped and resized to a 64 $\times$ 64 resolution. In addition, common data augmentation methods, including zooming, shifting, shearing, and flipping were randomly activated. The Adam optimization method \cite{adam} was used for optimizing the network parameters over 90 epochs, with a batch size of 128. The learning rate was set to 0.002 initially and reduced by a factor 0.1 every 30 epochs. Regarding loss hyperparameters, we used the standard configuration proposed in \cite{qd}: confidence interval set to 0.95, soften factor set to 160.0 and $\lambda=15.0$. For \MethodName we used the same setting, with $\beta=0.1$. Since there was no predefined test set for this dataset, we employed a 5-fold cross validation: In each split, we used 20\% as the test set. Additionally, 20\% of the train set was designated as the validation set. Best model obtained by minimizing the validation loss. In QD and \MethodName, we normalized ages to zero mean and unit variance.

\textbf{RSNA pediatric bone age dataset} \footnote{\url{https://www.kaggle.com/kmader/rsna-bone-age}}.\ \
For the RSNA dataset, we used the well-known VGG-16 architecture \cite{vgg} as a base model, with weights pre-trained on ImageNet. On top of this architecture, we added batch normalization \cite{batch_norm}, attention mechanism with two CNN layers of 64 and 16 neurons each, two average pooling layers, dropout \cite{dropout} with a 0.25 probability, and a fully connected layer with 1024 neurons. The activation function for the CNN layers was ReLU \cite{relu}, and we used ELU for the fully connected layer.  For the PIs last layer, we used biases of $[2.0, -2.0]$, for the upper and lower bound initialization, respectively. We used standard data augmentation consisting of horizontal flips, vertical and horizontal shifts, and rotations. In addition, we normalized targets to zero mean and unit variance. To reduce computational costs, we downscaled input images to  384 $\times$ 384 pixels. The network was optimized using Adam optimizer \cite{adam}, with an initial learning rate of 0.01 which was reduced when the validation loss has stopped improving over 10 epochs. We trained the network for 50 epochs using batch size of 100.  For our loss hyperparameters, we used the standard configuration like proposed in \cite{qd}: confidence interval set to 0.95, soften factor set to 160.0 and $\lambda=15.0$. For \MethodName, we used the same setting, with $\beta=0.5$.

\newpage
\section{UCI Analysis full results}
\label{sec:appendix_ablation}
In Table \ref{table:appendix_ablation_uci_table_pis} we present the full results of our ablation studies, including PICP, for the ablation variants.

\begin{table*}[ht]
\centering
\caption{Ablation analysis, comparing PICP, MPIW and RMSE. The guidelines for selecting best results are as those used in Table \ref{table:uci_table}.}
\label{table:appendix_ablation_uci_table_pis}
\resizebox{\textwidth}{!}{
  \begin{tabular}{l|ccc|ccc|ccc}
    \toprule
    \multirow{2}{*}{} &
      \multicolumn{3}{c|}{PICP} &
      \multicolumn{3}{c|}{MPIW} &
      \multicolumn{3}{c}{RMSE}  
      \\
      Datasets & {\OnlyRmse} & {\Vhalf} & {\MethodName} & {\OnlyRmse} & {\Vhalf} & {\MethodName} & {\OnlyRmse} & {\Vhalf} & {\MethodName} \\
      \midrule
    Boston & \textbf{0.93 $\pm$ 0.01} & \textbf{0.93 $\pm$ 0.01} & \textbf{0.93 $\pm$ 0.01} & \textbf{1.09 $\pm$ 0.02} & 1.15 $\pm$ 0.02 & \textbf{1.09 $\pm$ 0.01 } & 3.21 $\pm$ 0.24 & 3.39 $\pm$ 0.27 & \textbf{3.13 $\pm$ 0.21} \\
    Concrete & \textbf{0.93 $\pm$ 0.01} & \textbf{0.93 $\pm$ 0.01} & \textbf{0.93 $\pm$ 0.01} & \textbf{1.02 $\pm$ 0.01} & 1.07 $\pm$ 0.01 & \textbf{1.02 $\pm$ 0.01} & 5.55 $\pm$ 0.11 & 5.73 $\pm$ 0.10 & \textbf{5.43 $\pm$ 0.13} \\
    Energy & \textbf{0.97 $\pm$ 0.01} & \textbf{0.97 $\pm$ 0.00} & \textbf{0.97 $\pm$ 0.00} & \textbf{0.42 $\pm$ 0.01} & 0.45 $\pm$ 0.01 & \textbf{0.42 $\pm$ 0.01} & 2.16 $\pm$ 0.04 & 2.27 $\pm$ 0.04 & \textbf{1.65 $\pm$ 0.03} \\
    Kin8nm & \textbf{0.96 $\pm$ 0.00} & \textbf{0.96 $\pm$ 0.00} & \textbf{0.96 $\pm$ 0.00} & 1.13 $\pm$ 0.00 & 1.17 $\pm$ 0.00 & \textbf{1.10 $\pm$ 0.00} & 0.08 $\pm$ 0.00 & 0.08 $\pm$ 0.00 & \textbf{0.07 $\pm$ 0.00 } \\
    Naval & \textbf{0.98 $\pm$ 0.00} & \textbf{0.98 $\pm$ 0.00} & \textbf{0.98 $\pm$ 0.00} & \textbf{0.24 $\pm$ 0.00} & 0.30 $\pm$ 0.02 & \textbf{0.24 $\pm$ 0.00} & \textbf{0.00 $\pm$ 0.00} & \textbf{0.00 $\pm$ 0.00} & \textbf{0.00 $\pm$ 0.00}\\
    Power & \textbf{0.96 $\pm$ 0.00} & \textbf{0.96 $\pm$ 0.00} & \textbf{0.96 $\pm$ 0.00} & \textbf{0.86 $\pm$ 0.00} & \textbf{0.86 $\pm$ 0.00} & \textbf{0.86 $\pm$ 0.00} & 4.13 $\pm$ 0.04 & 4.15 $\pm$ 0.04 & \textbf{4.08 $\pm$ 0.04} \\
    Protein & \textbf{0.95 $\pm$ 0.00 }& \textbf{0.95 $\pm$ 0.00} & \textbf{0.95 $\pm$ 0.00} & \textbf{2.25 $\pm$ 0.01} & 2.27 $\pm$ 0.01 & \textbf{2.26 $\pm$ 0.01} & 4.78 $\pm$ 0.02 & 4.99 $\pm$ 0.01 & \textbf{4.35 $\pm$ 0.02} \\
    Wine & \textbf{0.91 $\pm$ 0.01} & \textbf{0.91 $\pm$ 0.01} & \textbf{0.91 $\pm$ 0.01} & 2.24 $\pm$ 0.01 & 2.23 $\pm$ 0.01 & \textbf{2.22 $\pm$ 0.01}  & 0.64 $\pm$ 0.01 & 0.67 $\pm$ 0.01 & \textbf{0.63 $\pm$ 0.01 }\\
    Yacht & \textbf{0.95 $\pm$ 0.01} & \textbf{0.95 $\pm$ 0.01} & \textbf{0.95 $\pm$ 0.01} & 0.18 $\pm$ 0.00 & 0.19 $\pm$ 0.01 & \textbf{0.17 $\pm$ 0.00} & 0.99 $\pm$ 0.07 & 1.15 $\pm$ 0.08 & \textbf{0.98 $\pm$ 0.07} \\
    MSD & \textbf{0.95 $\pm$ NA} & \textbf{0.95 $\pm$ NA} & \textbf{0.95 $\pm$ NA} & \textbf{2.42 $\pm$ NA} & 2.43 $\pm$ NA & \textbf{2.42 $\pm$ NA} & 9.10 $\pm$ NA & 9.25 $\pm$ NA &\textbf{ 8.93 $\pm$ NA} \\
    \bottomrule
  \end{tabular}
  }
\end{table*}

Additionally, we tested the Deep Ensembles baseline with an ensemble size of one. We expect the baseline to under-perform in this setting, and this expectation is supported by the results presented in Table \ref{table:de_one_uci_table}.

\begin{table*}[ht]
\centering
\caption{Results on regression benchmark UCI datasets comparing PICP, MPIW, and RMSE. The guidelines for selecting best results are as those used in Table \ref{table:uci_table}. Note: the results for DE are presented for ensemble size of one, denotes as \textit{DE-One}.}
\label{table:de_one_uci_table}
\resizebox{\textwidth}{!}{
  \begin{tabular}{l|ccc|ccc|ccc}
    \toprule
    \multirow{2}{*}{} &
      \multicolumn{3}{c|}{PICP} &
      \multicolumn{3}{c|}{MPIW} &
      \multicolumn{3}{c}{RMSE}  
      \\
      Datasets & {DE-One} & {QD} & {\MethodName} & {DE-One} & {QD} & {\MethodName} & {DE-One} & {QD} & {\MethodName} \\
      \midrule  
    Boston & 0.76 $\pm$ 0.01 & \textbf{0.93 $\pm$ 0.01} & \textbf{0.93 $\pm$ 0.01} & 0.75 $\pm$ 0.02 & 1.15 $\pm$ 0.02 & 1.09 $\pm$ 0.01 & \textbf{3.11 $\pm$ 0.21} & 3.39 $\pm$ 0.26 & 3.13 $\pm$ 0.21 \\
    Concrete & 0.87 $\pm$ 0.01 & \textbf{0.93 $\pm$ 0.01} & \textbf{0.93 $\pm$ 0.01} & 0.93 $\pm$ 0.02 & 1.08 $\pm$ 0.01 & 1.02 $\pm$ 0.01 & 5.52 $\pm$ 0.13 & 5.88 $\pm$ 0.10 & \textbf{5.43 $\pm$ 0.13} \\
    Energy & 0.93 $\pm$ 0.01 & \textbf{0.97 $\pm$ 0.01} & \textbf{0.97 $\pm$ 0.00} & 0.43 $\pm$ 0.02 & 0.45 $\pm$ 0.01 & \textbf{0.42 $\pm$ 0.01} & 1.72 $\pm$ 0.06 & 2.28 $\pm$ 0.04 &\textbf{1.65 $\pm$ 0.05 } \\
    Kin8nm & 0.93 $\pm$ 0.00 & \textbf{0.96 $\pm$ 0.00} & \textbf{0.96 $\pm$ 0.00} & 1.06 $\pm$ 0.01 & 1.18 $\pm$ 0.00 & \textbf{1.10 $\pm$ 0.00 } & 0.08 $\pm$ 0.00 & 0.08 $\pm$ 0.00 & \textbf{0.07 $\pm$ 0.00} \\
    Naval & 0.94 $\pm$ 0.00 & \textbf{0.97 $\pm$ 0.00} & \textbf{0.98 $\pm$ 0.00} & 0.25 $\pm$ 0.01 & 0.27 $\pm$ 0.00 & \textbf{0.24 $\pm$ 0.00 } & \textbf{0.00 $\pm$ 0.00} & \textbf{0.00 $\pm$ 0.00} & \textbf{0.00 $\pm$ 0.00} \\
    Power & \textbf{0.96 $\pm$ 0. 00} & \textbf{0.96 $\pm$ 0.00} & \textbf{0.96 $\pm$ 0.00} & 0.90 $\pm$ 0.00 & \textbf{0.86 $\pm$ 0.00} & \textbf{0.86 $\pm$ 0.00 } & \textbf{4.01 $\pm$ 0.04} & 4.14 $\pm$ 0.04 & 4.08 $\pm$ 0.04 \\
    Protein & \textbf{0.95 $\pm$ 0.00 }& \textbf{0.95 $\pm$ 0.00} & \textbf{0.95 $\pm$ 0.00} & 2.64 $\pm$ 0.01 & \textbf{2.27 $\pm$ 0.01} & \textbf{2.26 $\pm$ 0.01} & 4.43 $\pm$ 0.02 & 4.99 $\pm$ 0.02 & \textbf{4.35 $\pm$ 0.02}\\
    Wine & 0.86 $\pm$ 0.01 & \textbf{0.91 $\pm$ 0.01} & \textbf{0.91 $\pm$ 0.01} & 2.34 $\pm$ 0.02 & \textbf{2.24 $\pm$ 0.02} & \textbf{2.22 $\pm$ 0.01  } & \textbf{0.64 $\pm$ 0.01} & 0.67 $\pm$ 0.01 & \textbf{0.63 $\pm$ 0.01} \\
    Yacht & \textbf{0.95 $\pm$ 0.01} & \textbf{0.95 $\pm$ 0.01} & \textbf{0.95 $\pm$ 0.01} & 0.26 $\pm$ 0.02 & 0.18 $\pm$ 0.00 & \textbf{0.17 $\pm$ 0.00 } & 1.42 $\pm$ 0.07 & 1.10 $\pm$ 0.06 & \textbf{0.98 $\pm$ 0.07} \\
    MSD & \textbf{0.95 $\pm$ NA} & \textbf{0.95 $\pm$ NA} & \textbf{0.95 $\pm$ NA} & 2.86 $\pm$ NA & 2.45 $\pm$ NA & \textbf{2.42 $\pm$ NA} & 9.03 $\pm$ NA & 9.30 $\pm$ NA & \textbf{8.93 $\pm$ NA} \\ 
    \bottomrule
  \end{tabular}
  }
\end{table*}

\begin{table}[ht]
\centering
\RawFloats
\caption{an extension of Table \ref{table:uci_table} evaluating MPIW with PICP for all our baselines.}
\label{table:appendix_uci_table_pis}
\resizebox{\textwidth}{!}{
  \begin{tabular}{l|cccc|cccc}
    \toprule
    \multirow{2}{*}{} &
      \multicolumn{4}{c|}{\textbf{PICP}} &
      \multicolumn{4}{c}{\textbf{MPIW}} 
      \\
      Datasets & {DE} & {QD} & {QD+} & {\MethodName} & {DE} & {QD} & {QD+} & {\MethodName} \\
      \midrule
    Boston & 0.87 $\pm$ 0.01 & \textbf{0.93 $\pm$ 0.01}  & \textbf{1.00 $\pm$ 0.01} & \textbf{0.93 $\pm$ 0.01} &  0.87 $\pm$ 0.03 & 1.15 $\pm$ 0.02 & 4.98 $\pm$ 0.81 & 1.09 $\pm$ 0.01 \\
    Concrete & 0.92 $\pm$ 0.01 & \textbf{0.93 $\pm$ 0.01} & \textbf{1.00 $\pm$ 0.01} & \textbf{0.93 $\pm$ 0.01} & 1.01 $\pm$ 0.02 & 1.08 $\pm$ 0.01 & 3.43 $\pm$ 0.26 & 1.02 $\pm$ 0.01 \\
    Energy & \textbf{0.99 $\pm$ 0.00} & \textbf{0.97 $\pm$ 0.01} & \textbf{0.99 $\pm$ 0.01} & \textbf{0.97 $\pm$ 0.00} & 0.49 $\pm$ 0.01 & 0.45 $\pm$ 0.01 & 1.52 $\pm$ 0.18 & \textbf{0.42 $\pm$ 0.01} \\
    Kin8nm & \textbf{0.97 $\pm$ 0.00} & \textbf{0.96 $\pm$ 0.00} & \textbf{0.99 $\pm$ 0.00} & \textbf{0.96 $\pm$ 0.00} & 1.14 $\pm$ 0.01 & 1.18 $\pm$ 0.00 & 2.64 $\pm$ 0.19 &\textbf{1.10 $\pm$ 0.00 } \\
    Naval & \textbf{0.98 $\pm$ 0.00} & \textbf{0.97 $\pm$ 0.00} & \textbf{0.99 $\pm$ 0.01} & \textbf{0.98 $\pm$ 0.00} & 0.31 $\pm$ 0.01 & 0.27 $\pm$ 0.00 & 2.85 $\pm$ 0.15 & \textbf{0.24 $\pm$ 0.00 }\\
    Power & \textbf{0.96 $\pm$ 0. 00} & \textbf{0.96 $\pm$ 0.00} & \textbf{0.99 $\pm$ 0.00} & \textbf{0.96 $\pm$ 0.00} & 0.91 $\pm$ 0.00 & \textbf{0.86 $\pm$ 0.00} & 1.51 $\pm$ 0.06 & \textbf{0.86 $\pm$ 0.00 }\\
    Protein & \textbf{0.96 $\pm$ 0.00 }& \textbf{0.95 $\pm$ 0.00} & \textbf{0.99 $\pm$ 0.00} & \textbf{0.95 $\pm$ 0.00} & 2.68 $\pm$ 0.01 & \textbf{2.27 $\pm$ 0.01} & 3.30 $\pm$ 0.03 & \textbf{2.26 $\pm$ 0.01} \\
    Wine & 0.90 $\pm$ 0.01 & \textbf{0.91 $\pm$ 0.01} & \textbf{1.00 $\pm$ 0.00} & \textbf{0.91 $\pm$ 0.01} & 2.50 $\pm$ 0.02 & \textbf{2.24 $\pm$ 0.02} & 4.89 $\pm$ 0.12 & \textbf{2.22 $\pm$ 0.01  }\\
    Yacht & \textbf{0.98 $\pm$ 0.01} & \textbf{0.95 $\pm$ 0.01} & \textbf{0.97 $\pm$ 0.03} & \textbf{0.95 $\pm$ 0.01} & 0.33 $\pm$ 0.02 & 0.18 $\pm$ 0.00 & 1.25 $\pm$ 0.23 & \textbf{0.17 $\pm$ 0.00 } \\
    MSD & \textbf{0.95 $\pm$ NA} & \textbf{0.95 $\pm$ NA}& \textbf{0.99 $\pm$ NA} & \textbf{0.95 $\pm$ NA} & 2.91 $\pm$ NA & 2.45 $\pm$ NA & 4.25 $\pm$ NA & \textbf{2.42 $\pm$ NA} \\
    \bottomrule
  \end{tabular}
  }
\end{table}

\section{Evaluation under Dataset Shift}
\label{appendix:ds_shift}
Dataset shift is a challenging problem in which the  dataset composition and/or distribution  changes over time. This scenario poses significant challenges to the application of machine learning algorithms, because these models tend to assume similar distributions between their train and test sets. Dataset shift scenarios are of particular interest to uncertainty modeling  \cite{can_you_trust}, as it enables researchers to evaluate algorithms' robustness and adaptability. We now evaluate \MethodName's ability to adjust to this challenging scenario.

\begin{figure}[ht]
\centering
\includegraphics[width=\textwidth]{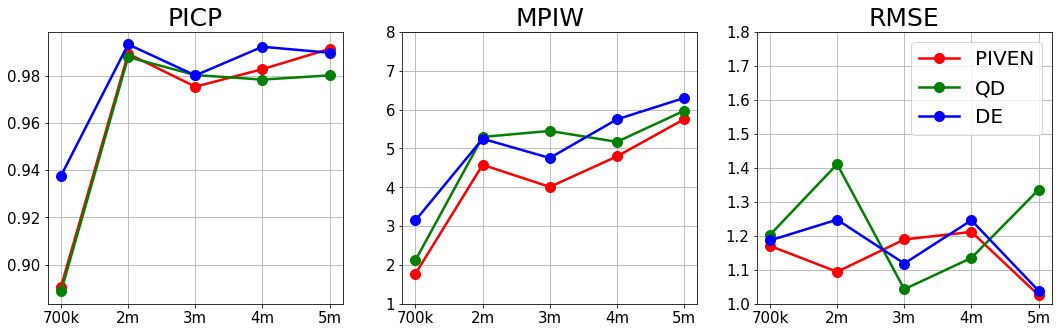}
\caption{Compassion between \MethodName and the baselines in terms of PICP (left), MPIW (center) and RMSE (right) for $\alpha=0.05$.}
\label{fig:flights_comparsion} 
\end{figure}

For our evaluation, we chose the Flight Delays dataset \cite{flights_ds}, which is known to contain dataset shift. We train \MethodName and our baselines on the first 700K data points and test on the next 100K test points at 5 different starting points: 700K, 2M (million), 3M, 4M and 5M. This experimental setting \textit{fully replicates} (both in terms of dataset splits and the neural architectures used in the experiments) the one presented in \cite{nips_flights}. The dataset is ordered chronologically throughout the year 2008, and it is evident that the dataset shift increases over time. Our experimental results -- using the PICP, MPIW and RMSE metrics -- are presented in Figure \ref{fig:flights_comparsion}. As in our other experiments, we present the results for a confidence level of 95\% (i.e. $\alpha=0.05$).

The results show that \MethodName outperforms the baselines in MPIW metric (a result consistent with our other experiments) while achieving the desried coverage rate. Interestingly, our approach is also competitive in terms of the RMSE metric, although the value fluctuations make it difficult to reach a clear conclusion. Neither approach reached the desired PICP level in the first test set, but all were able to do so in consequent experiments. DE was able to achieve slightly higher PICP rates than the other two methods, but its PI width were larger than the other two methods. In contrast, \MethodName manages to achieve the desired PICP levels while producing smaller intervals.
It should be noted that the dataset exhibits seasonal effect between 2M and 3M samples, which causes variance in performance. This phenomenon is described in \cite{nips_flights}.
We hypothesize that the reason \MethodName outperforms DE in terms of PI width is due to the fact that the former aims to optimize MPIW, whereas DE aims to optimize for negative log-likelihood \cite{deep_ensemble}. Additionally, the Flight Delays dataset is known to have a non-Gaussian distribution \cite{flights_non_gaussian} which invalidates the basic assumption applied by DE.

\section{Full Results - PI Performance as a Function of Coverage}
\label{sec:appendix_alpha_full_results}
\begin{figure}[ht]
\centering
\includegraphics[width=\textwidth]{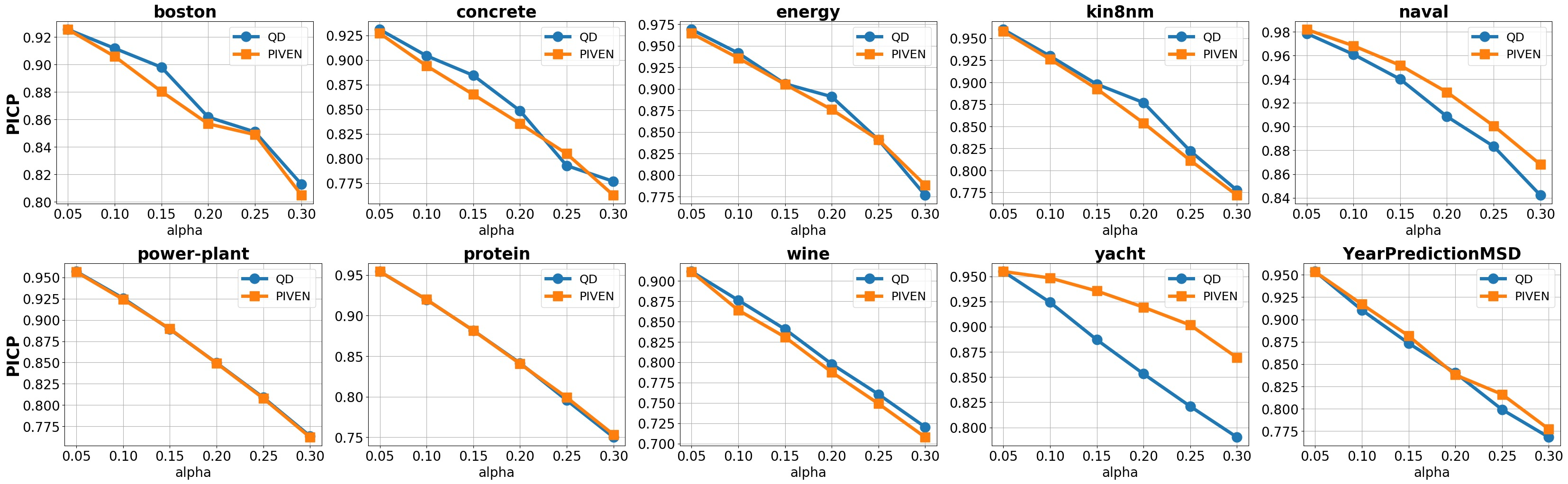}
\caption{comparing PICP between QD and \MethodName over different values of alpha.}
\label{fig:alpha_picp} 
\end{figure}

\begin{figure}[ht]
\centering
\includegraphics[width=\textwidth]{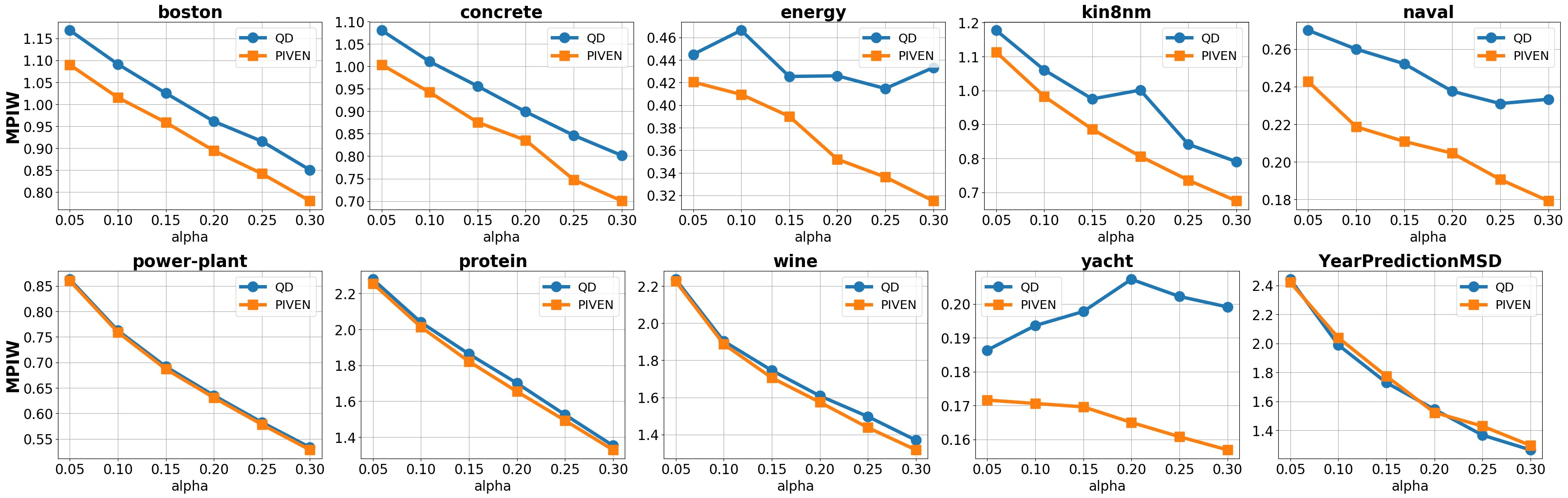}
\caption{comparing MPIW between QD and \MethodName over different values of alpha.}
\label{fig:alpha_mpiw} 
\end{figure}

\begin{figure}[ht]
\centering
\includegraphics[width=\textwidth]{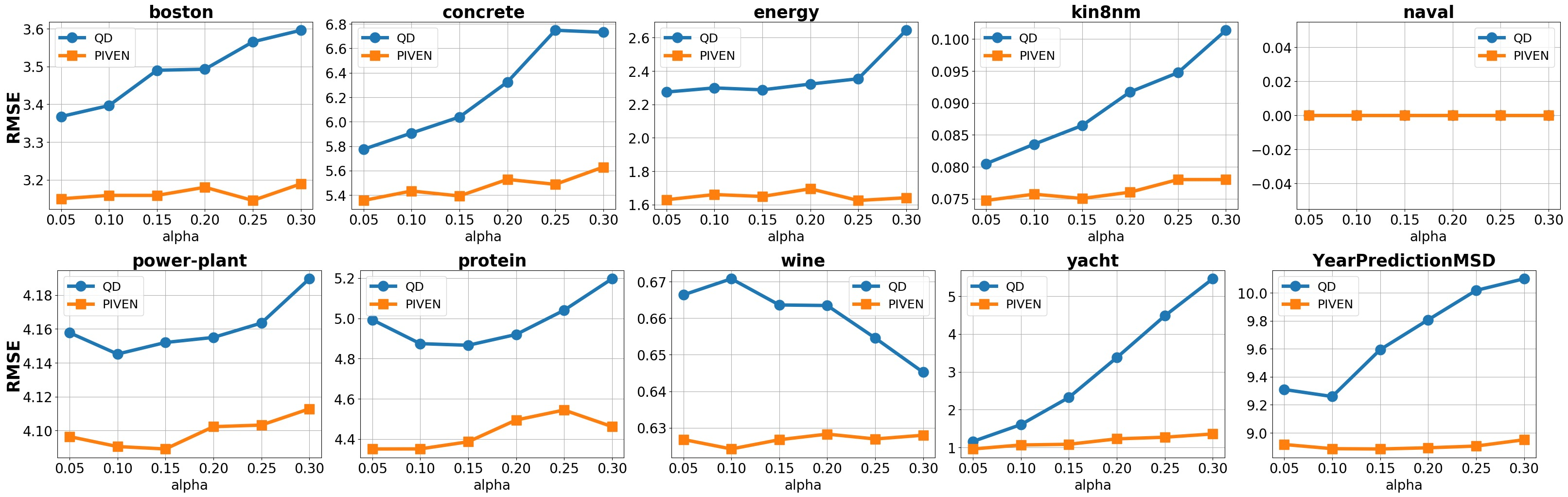}
\caption{comparing RMSE between QD and \MethodName over different values of alpha.}
\label{fig:alpha_rmse} 
\end{figure}

We argue that QD only focuses on the fraction $c$ of the training set which was successfully captured by the PI. In this experiment we study the effect of changes in the coverage levels (represented by $\alpha$) on the width of the PI produced both by \MethodName and QD. Additionally, we examine the affect of \(\alpha\) on the value prediction. In Figures \ref{fig:alpha_picp}, \ref{fig:alpha_mpiw} and \ref{fig:alpha_rmse}, we present our results -- measured by PICP, MPIW and RMSE -- on all UCI datasets. As expected, \MethodName consistently outperforms QD in the above metrics. 

\textbf{PI Coverage.} \ \  Both methods are able to achieve the coverage desired by \(\alpha\), which decreases as \(\alpha\) increases, as expected.

\newpage
\textbf{PI Width.} \ \ QD's inability to consider points outside the PI clearly degrades its performance. \MethodName, on the other hand, is able to reduce MPIW consistently over all datasets. 

\textbf{Value prediction accuracy.} \ \ Due to the way QD performs its value prediction (i.e., middle of the PI), the value prediction is not able to improve, and in fact degrades. Contrarily, \MethodName's robustness enables it to generally maintain (with a slight decrease) its performance levels.

\section{IMDB age estimation training process and robustness to outliers}
\label{sec:imdb_training_process}

\subsection{Training process}
\begin{figure}[ht]
\resizebox{\textwidth}{!}{
\begin{tcbitemize}[
    raster columns=2,
    raster halign=center,
    raster every box/.style={blankest}
    ]
\mysubfig{MAE NN}{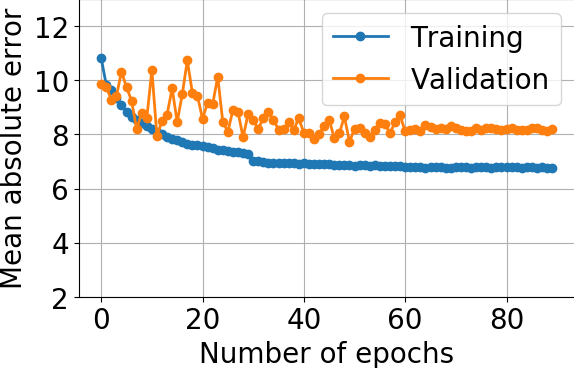}
\mysubfig{MAE QD}{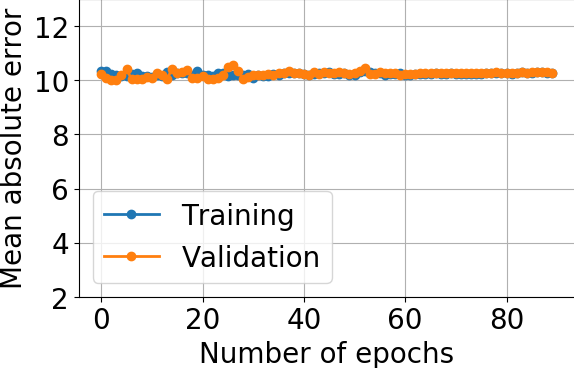}
\mysubfig{MAE \MethodName}{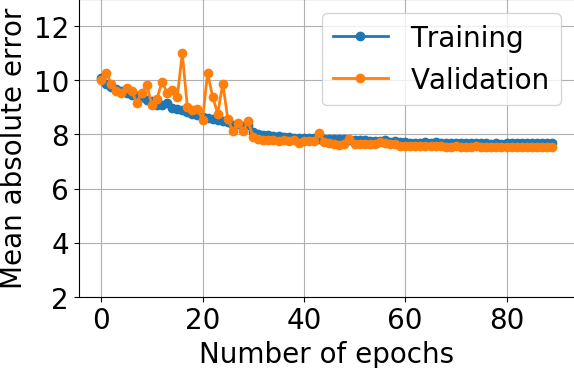}
\mysubfig{MAE validation errors}{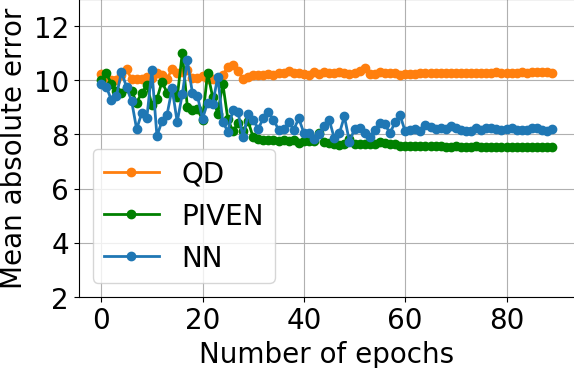}
\end{tcbitemize}
\caption{Comparison of MAE metric in the training process. We observe that the values for QD (b) do not improve, which is expected since QD does not consider value prediction in its loss function. Moreover, we note that NN suffers from overfitting, given that the validation error is greater than the training error after convergence. This phenomena do not affect \MethodName, thus providing an indication of its robustness.}
\label{fig:imdb_training_mae}
}
\end{figure}

In the following figures we present comparisons of the training progression for \MethodName, QD and NN on the MAE, PICP and MPIW evaluation metrics. We used 80\% of images as the training set while the remaining 20\% were used as the validation set (we did not define a test set as we were only interested in analyzing the progression of the training). For the MAE metric, presented in Figure \ref{fig:imdb_training_mae}, we observe that the values for QD did not improve. This is to be expected since QD does not consider this goal in its training process (i.e., loss function). This result further strengthens our argument that selecting the middle of the interval is often a sub-optimal strategy for value prediction. For the remaining two approaches -- NN and \MethodName -- we note that NN suffers from overfitting, given that the validation error is greater than training error after convergence. This phenomena does not happen in \MethodName which indicates robustness, a result which further supports our conclusions regarding the method's robustness. 

\begin{figure}[ht]
\resizebox{0.8\textwidth}{!}{
\begin{tcbitemize}[
    raster columns=2,
    raster halign=center,
    raster every box/.style={blankest}
    ]
\mysubfig{MPIW QD}{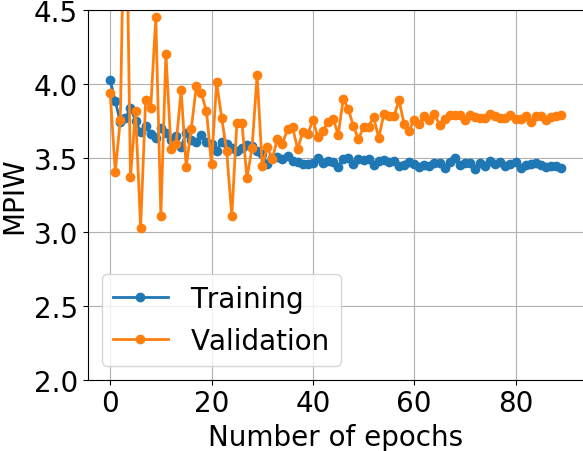}
\mysubfig{MPIW \MethodName}{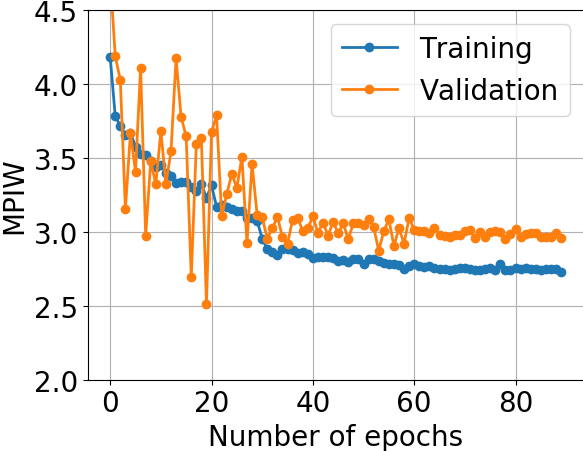}
\mysubfig{MPIW validation}{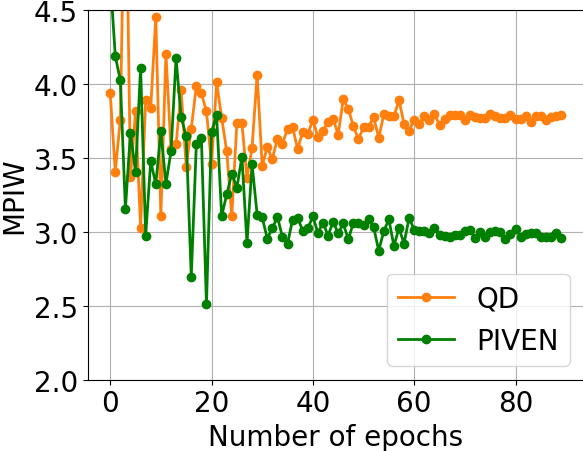}
\end{tcbitemize}
\caption{Comparison of the MPIW metric between QD and \MethodName in the training process. As can be seen, \MethodName significantly improves over QD, and has a smaller gap between training and validation errors.}
\label{fig:imdb_training_mpiw}
}
\end{figure}

\begin{figure}[ht]
\resizebox{0.8\textwidth}{!}{
\begin{tcbitemize}[
    raster columns=2,
    raster halign=center,
    raster every box/.style={blankest}
    ]
\mysubfig{PICP QD}{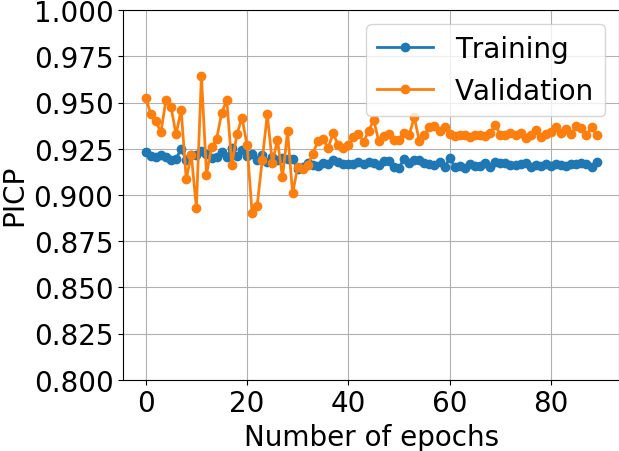}
\mysubfig{PICP \MethodName}{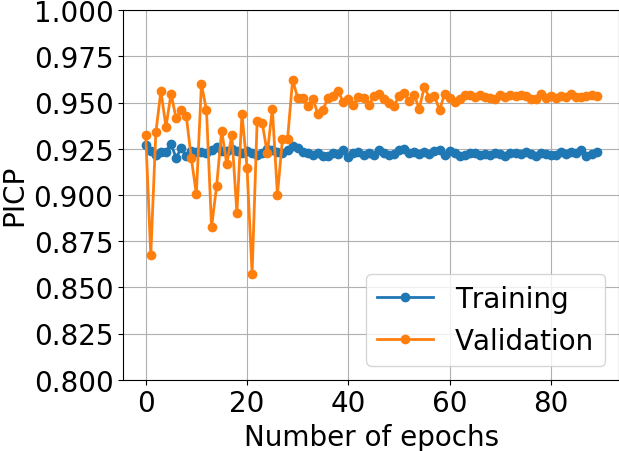}
\mysubfig{PICP validation}{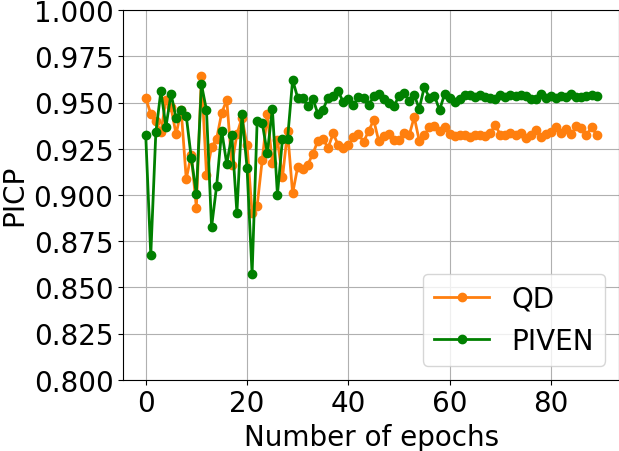}
\end{tcbitemize}
\caption{Comparison of PICP metric between QD and \MethodName in the training process. \MethodName achieves higher coverage when two methods converges.}
\label{fig:imdb_training_picp}
}
\end{figure}

For the MPIW metric (Figures \ref{fig:imdb_training_mpiw}), \MethodName presents better performance both for the validation and train sets compared to QD. Moreover, we observe a smaller gap between the error produced by \MethodName for the two sets -- validation and training -- which indicates that \MethodName enjoys greater robustness and an ability to not overfit to a subset of the data. Our analysis also shows that for the PICP metric (Figure \ref{fig:imdb_training_picp}), \MethodName achieves higher levels of coverage.

\newpage
\subsection{Robustness to outliers}
\begin{figure}[ht]
    \RawFloats
    \centering
    \begin{minipage}{0.45\textwidth}
        \centering
        \includegraphics[width=0.75\textwidth]{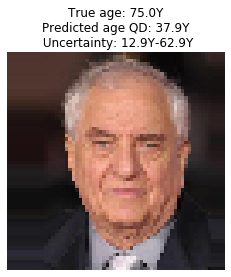} 
    \end{minipage}
    \begin{minipage}{0.45\textwidth}
        \centering
        \includegraphics[width=0.75\textwidth]{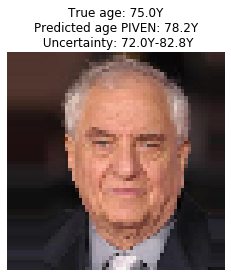}  
    \end{minipage}
    \begin{minipage}{0.45\textwidth}
        \centering
        \includegraphics[width=0.75\textwidth]{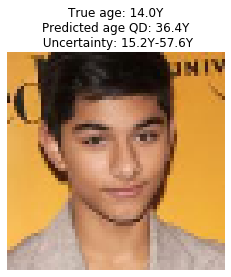}  
    \end{minipage}
    \begin{minipage}{0.45\textwidth}
        \centering
        \includegraphics[width=0.75\textwidth]{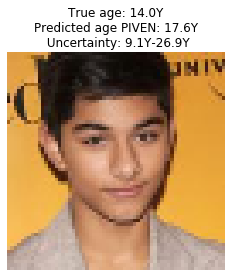}  
    \end{minipage}
    \begin{minipage}{0.45\textwidth}
        \centering
        \includegraphics[width=0.75\textwidth]{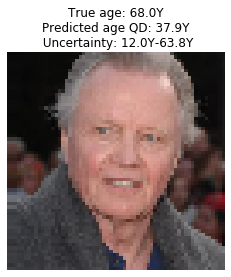}
    \end{minipage}
    \begin{minipage}{0.45\textwidth}
        \centering
        \includegraphics[width=0.75\textwidth]{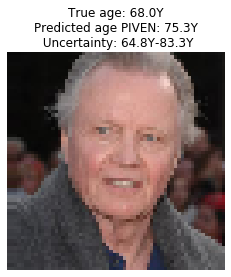}  
    \end{minipage}
    \caption{The predictions produced for outliers (i.e., very young/old individuals) by both \MethodName and QD for the IMDB age estimation dataset. The results for QD are on the left, results for \MethodName on the right.}
    \label{fig:imdb_outliers}
\end{figure}

Since \MethodName is capable of learning from the entire dataset while QD learns only from data points which were captured by the PI, it is reasonable to expect that the former will outperform the latter when coping with outliers. In the IMDB age estimation dataset, we can consider images with very high or very low age as outliers. Our analysis shows that for this subset of cases, there is a large gap in performance between \MethodName and QD. In Figure \ref{fig:imdb_outliers} we provide several images of very young/old individuals and the results returned by the two methods. We can observe that \MethodName copes with these outliers significantly better.

\end{document}